\documentclass[journal]{IEEEtran}

\ifCLASSINFOpdf
\else
\fi
\usepackage{cite}
\usepackage{amsmath,amssymb,amsfonts}
\usepackage{algorithm, algorithmic}
\usepackage{graphicx}
\usepackage{textcomp}
\usepackage{xcolor}
\usepackage{animate}
\usepackage{multirow}
\usepackage{tabularx}
\usepackage{booktabs}
\usepackage[normalem]{ulem}
\newcommand{\tabincell}[2]{\begin{tabular}{@{}#1@{}}#2\end{tabular}}

\begin{document}
%
\title{Semantic Neighborhood-Aware Deep Facial Expression Recognition}
%
%
%

\author{Yongjian Fu, Xintian Wu, Xi Li*, Zhijie Pan, Daxin Luo
\thanks{This work is in part supported by key scientific technological innovation research project by Ministry of Education, Zhejiang Provincial Natural Science Foundation of China under Grant LR19F020004, the National Natural Science Foundation of China under Grant 61751209, Zhejiang University K.P.Chao's High Technology Development Foundation.
The authors would like to thank Songyuan Li, Hui Wang, Hanbin Zhao and Xuewei Li for their valuable comments and suggestions.}
\thanks{Y. Fu, X. Wu are with College of Computer Science, Zhejiang University, Hangzhou 310027, China. (e-mail:
        {yjfu, hsintien@zju.edu.cn})}%
\thanks{X. Li* (corresponding author)is with the College of Computer Science and Technology, Zhejiang University, Hangzhou 310027, China.(e-mail:
        {xilizju@zju.edu.cn})}%
\thanks{Z. Pan is with the College of Computer Science and Technology, Zhejiang University, Hangzhou, China. (e-mail:
        {zhijie\_pan@zju.edu.cn})}%
\thanks{D.Luo is with Noah's Ark Lab, Huawei Technologies. (e-mail:
        {luodaxin@huawei.com})}
\thanks{The first two authors contribute equally.}

}

\markboth{IEEE Transactions on Image Processing,~Vol.~XX, No.~X,~202X}%
{Fu \MakeLowercase{\textit{et al.}}: Semantic Neighborhood-Aware Deep Facial Expression Recognition}

\maketitle

\begin{abstract}
Different from many other attributes, facial expression can change in a continuous way, and therefore, a slight semantic change of input should also lead to the output fluctuation limited in a small scale.
This consistency is important.
However, current Facial Expression Recognition (FER) datasets may have the extreme imbalance problem, as well as the lack of data and the excessive amounts of noise, hindering this consistency and leading to a performance decreasing when testing.
In this paper, we not only consider the prediction accuracy on sample points, but also take the neighborhood smoothness of them into consideration, focusing on the stability of the output with respect to slight semantic perturbations of the input.
A novel method is proposed to formulate semantic perturbation and select unreliable samples during training, reducing the bad effect of them.
Experiments show the effectiveness of the proposed method and state-of-the-art results are reported, getting closer to an upper limit than the state-of-the-art methods by a factor of 30\% in AffectNet, the largest in-the-wild FER database by now.

\end{abstract}


%
\IEEEpeerreviewmaketitle

\section{Introduction}
\label{motivation}
The goal of facial expression recognition (FER) is to recognize the basic human emotions, viz. \emph{Anger}, \emph{Disgust}, \emph{Fear}, \emph{Happiness}, \emph{Sadness} and \emph{Surprise}\cite{basic}, from a human facial image or sequence.
A certain facial expression can change in a continuous way across human facial manifold while corresponding image remains semantically significative, which requires the prediction of the FER model to change along with it continuously as well.
An appropriate FER model therefore should satisfy this consistency, which means if a neighborhood semantic perturbation (small facial variation of the same expression), e.g. slightly raising of lips, is added to the input image, the fluctuation of output prediction result should also be stable and limited in a small scale.

\begin{figure}[tp]
\centering
\begin{minipage}{\linewidth}
\includegraphics[width=\linewidth]{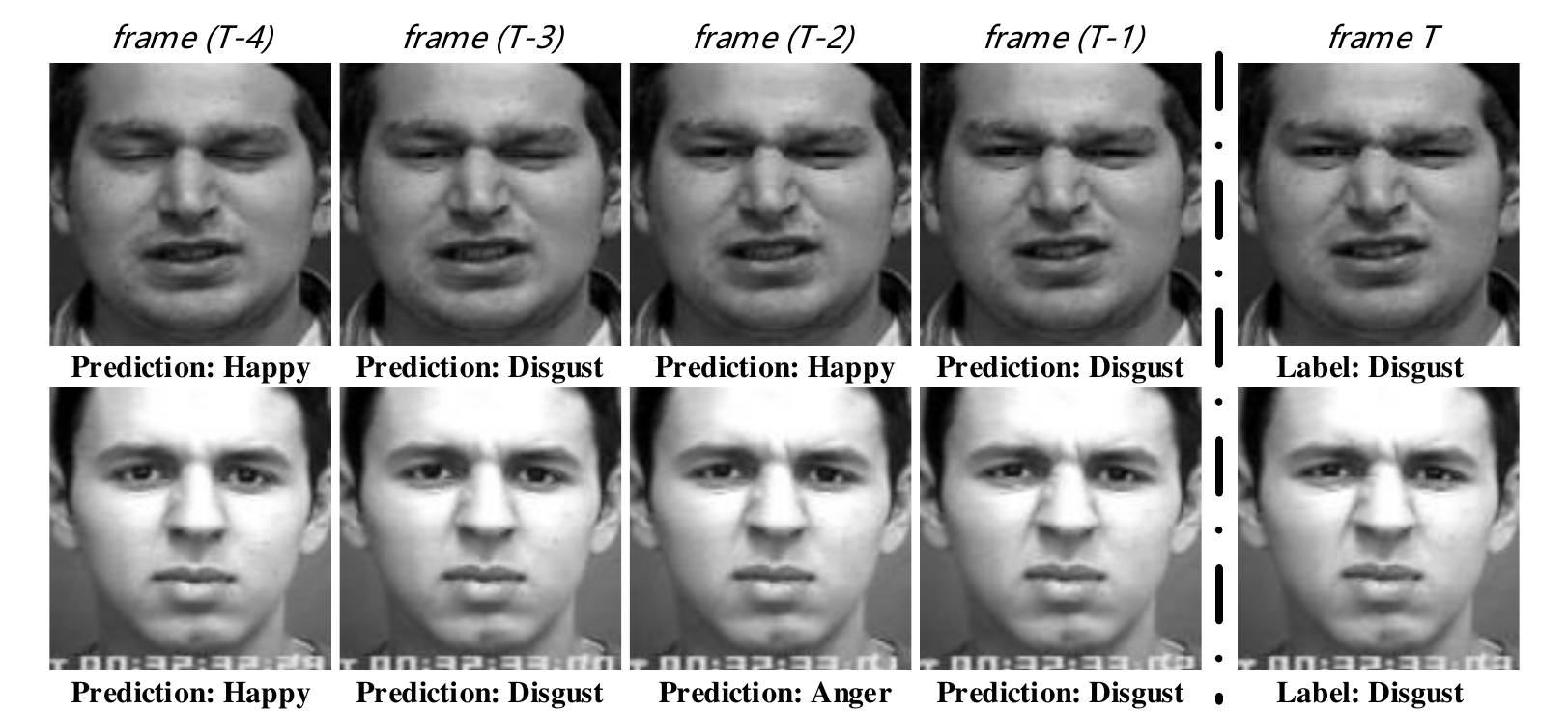}
\centerline{(a) Effect of Perturbation}
\end{minipage}
\begin{minipage}{\linewidth}
\includegraphics[width=\linewidth]{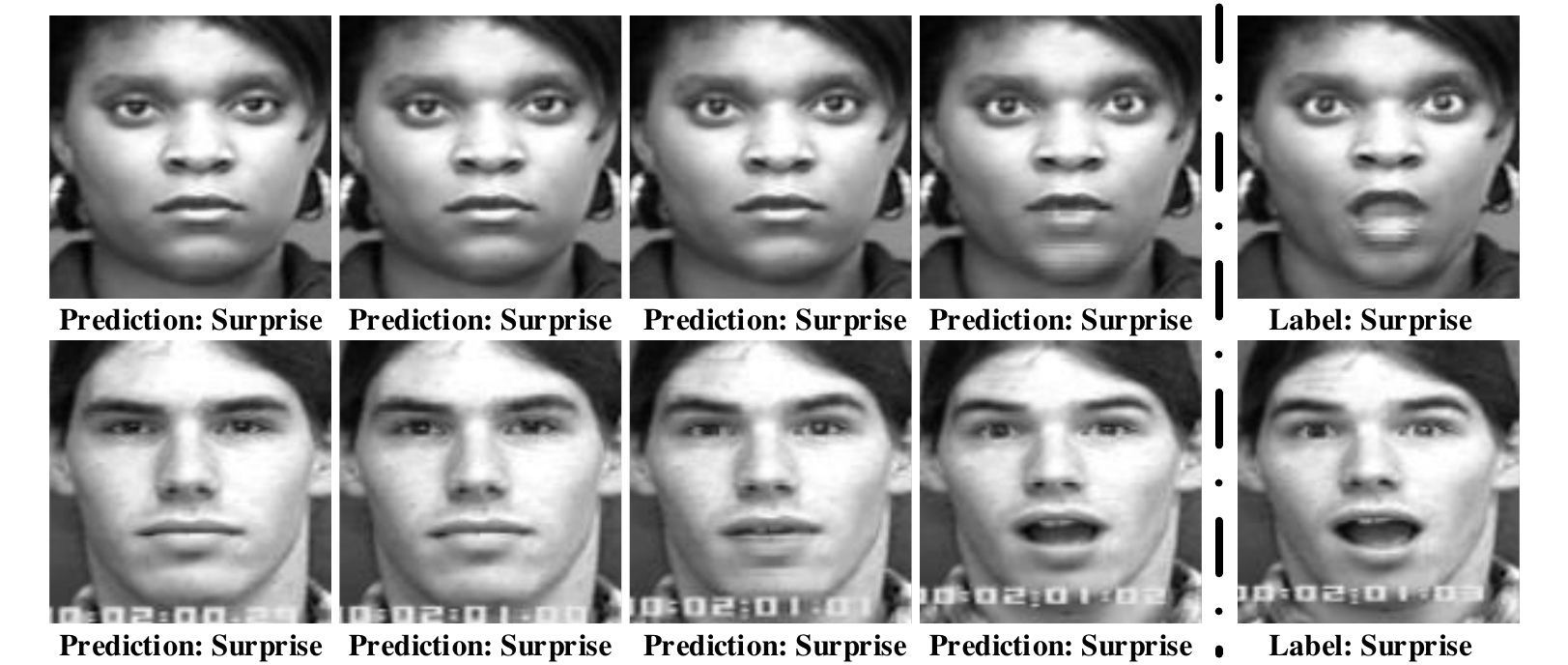}
\centerline{(b) Effect of Identity}
\end{minipage}
\caption{
Some prediction cases of our well-trained model. Images in the last column of each sub-figure are the last frame of an expression video, and we use them as training samples while those in the other columns (from the frame T-4 to T-1 in the video) are testing samples. (a): a small semantic change in input sample can lead to misclassifying of the CNN model. (b): the CNN model learns to recognize expressions by identity, thus neutral images are predicted as \emph{Surprise}. }
\label{overlearning}
\end{figure}

However, some characteristics of FER datasets hinder this consistency, especially when the widely-used deep learning fashion is applied.
Noisy data is a noticeable part in FER datasets because of the subjectivity in annotation \cite{affectnet}, \cite{trans-inconsis}, and, what is worse, some FER datasets are of relatively small scales \cite{ck+}, \cite{oulu} or have the extreme imbalance problem.
With the great fitting ability, a deep model can overfit some samples (including some noises, samples of minority classes, and even some normal samples) accurately while ignoring neighborhood smoothness and the output stability.
That is to say, a well-trained FER deep model is easy to get overlearned \cite{overlearning} and are vulnerable to disturbance, meaning that (1) a small semantic change of input sample may cause a non-negligible fluctuation on the prediction result, and (2) expressions can be recognized by statistically uncorrelated variables, e.g. identity, as shown in Fig~\ref{overlearning}.
Recent works mainly focus on the prediction accuracy on sample points, while the neighborhood smoothness and the consistency of input and output change are lack of due consideration, hindering the performance of prediction models.

The goal, in this paper, is to enhance this consistency, as well as the stability of the output w.r.t. slight semantic perturbation of the input, thereby improve the performance of FER model. Apart from the prediction accuracy of a single sample point, we take the model smoothness of the neighborhood around the given sample into consideration, supposing that:
\begin{enumerate}

\item Training with the unreliable samples, around which the function curve or the boundary of model is not smooth and stable, has a negative effect on the prediction model, increasing the risk of overlearning as well as hurting the performance.
\item If the function curve around a neighborhood of the given sample is not smooth and stable, the prediction results of the given sample and its semantic neighbor (the original given sample with a slight semantic change or small facial variation of the same expression) can be very inconsistent.

\end{enumerate}
Along with the assumptions, we can distinguish the unreliable samples (including noise samples) according to the neighborhood smoothness with the help of semantic neighbors, and thereby reduce the bad effect of those unreliable samples to enhance both the consistency and the performance.

\begin{figure*}[tp]
\centerline{\includegraphics[width=1\textwidth]{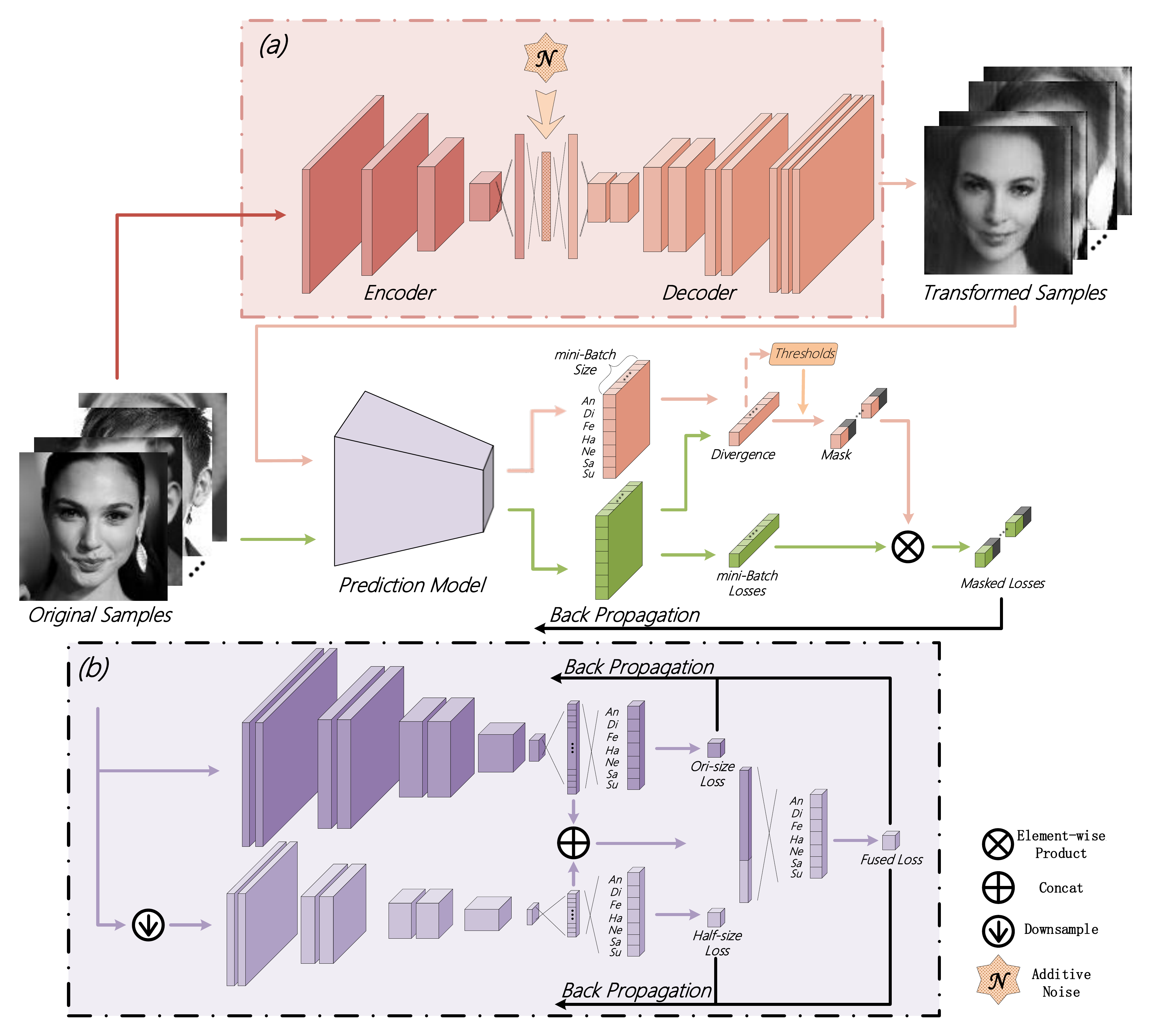}}
\caption{Overview of the proposed method. The mask computed with semantic neighbor and threshold achieves our method when training, which is described in Section \ref{DBDS}. (a) shows the architecture of our asymmetric autoencoder, with additive noise applied on the latent code to formulate the referred neighborhood semantic transformation, which is introduced in Section~\ref{GN}. (b) shows the structure of our prediction model, which is a two-scale network in fact. More details of the network setting can be found in TABLE~\ref{arch_ae} and TABLE~\ref{arch_pred}. An: Anger; Di: Disgust; Fe: Fear; Ne:Neutral; Ha: Happy; Sa: Sad; Su: Surprise.}
\label{overview}
\end{figure*}

We propose a novel method consisting of two stages to cope with this task, shown in Fig~\ref{overview}.
In the first stage, we conduct a neighborhood semantic transformation to synthesize semantic neighbors of a given sample. To tackle the tricky high-dimensional data, i.e. images, we project the sample into a semantic latent space to formulate the semantic perturbation on a relatively low-dimensional space rather than process it in pixel level directly.
In the second stage, we detect unreliable samples following the assumptions with the synthesized semantic neighbors. As we supposed, these samples have a negative effect on the prediction model when training and we therefore reduce their stimulation to the critic boundary.
Experiments show the significant improvement of performance, proving the effectiveness of the proposed method.

In summary, the contribution of this paper is three-fold. Firstly, we propose a novel optimization method to enhance the stability of the output w.r.t. slight semantic perturbations of the input, focusing on neighborhood smoothness of the deep model. Secondly, we design a whole framework consisting of neighborhood semantic transformation and semantic neighborhood-aware optimization strategy to implement the algorithm. Finally, experiments are conducted to prove the effectiveness of our method and we get closer to an upper limit than the state-of-the-art methods by a factor of 30\% in AffectNet \cite{affectnet}, which is the largest in-the-wild FER database by now.

The remainder of this paper is structured as follows. In the next section, we briefly review the development and some typical methods of expression recognition. Then we demonstrate a detailed discussion on our principle and introduce the pipeline of our algorithm part by part in Section~\ref{APP}. Section~\ref{EXP} provides the experimental results on AffectNet, CK+, and Oulu-CASIA. Conclusion and future works are finally stated in Section~\ref{CON}.

\section{RELATED WORK}

\subsection{Extracting Expression Feature}
Conventional facial expression recognition methods mainly focus on extracting effective and robust feature. Many handcrafted features have been designed and utilized in FER task over the years, such as LBP \cite{lbp}, HOG \cite{hog}, SIFT \cite{sift} and Haar \cite{haar} for static features extraction, as well as Cov3D \cite{cov3d}, MSR \cite{msr} and LBP-TOP \cite{lbptop} for 3D spatiotemporal features. Methods based on action unions \cite{au}, \cite{au2} and other mid-level features \cite{mid-feat1}, \cite{mid-feat2} are also proposed to bridge the gap between low-level features and high-level semantics. To make full use of the extracted features, researchers also take advantage of classifiers fusion algorithms on the decision-level to combine several single classification methods, e.g. SVM, DBN, and HMM \cite{au}, \cite{fuse}.

Recently a number of methods based on deep learning have been proposed to solve the difficult biometrics tasks \cite{VincenzoAge}, \cite{Biometrics}, and for FER task, they have also achieved state-of-the-art performance\cite{deep1}, \cite{deep2}. To obtain more pointed deep features, some works \cite{cnn1}, \cite{cnn2}, \cite{cnn3} improve the adaptability of models in FER datasets by designing architectures for this specific task, while others integrate deep models with prior facial knowledge to enhance the ability of emotion comprehension. Landmarks are used in \cite{prior-landmark}, \cite{prior-landmark2}, \cite{prior-landmark3} to focus more attention on critical details, and therefore the facial-geometric-aware features can be obtained naturally. In \cite{prior-au} researchers believe that expression can be decomposed into a batch of AUs, thus an AU-aware layer is designed to extract the feature of AUs combinations. In \cite{prior-depth}, depth information is used and Patterns of Oriented Motion Flow (POMF) are proposed to discretize the motion change. In \cite{prior-view}, facial orientation is taken into consideration to generate expression features which are invariance to the change of facial views.

\subsection{Solving Dataset Characteristic}
Datasets of FER may have their own problems.
Some FER datasets have relatively small scales \cite{ck+}, \cite{oulu} with only hundreds of samples in the datasets, while others may have the extreme imbalance problem. Besides, excessive amounts of noise may distribute in the dataset because of the subjectivity in annotation \cite{affectnet}, \cite{trans-inconsis}.
Therefore, methods aiming at the characteristic of datasets are proposed to remedy those flaws.

To solve the lack of facial samples, \cite{trans1}, \cite{trans2} pre-train the model on a low-resolution dataset with relatively more images\cite{FER}, which leads to a two-stage training procedure. Since facial recognition datasets typically contains a large amount of samples \cite{Learned2016Labeled}, \cite{MegaFace}, researchers propose to utilize these labeled facial data \cite{trans-f2e}, and a hint-based model compression method \cite{hint} is used to transfer the learned knowledge to a smaller model, which then will be fine-tuned on a target FER dataset. To make full use of scarce sequence datasets, \cite{trans-peak} proposes Peak Gradient Suppression (PGS) to guide the training of hard samples. Multiple datasets are merged in \cite{trans-meta} and a meta-learning method is conducted to keep the knowledge learned from each of them. Further, \cite{trans-inconsis} focuses mainly on the inconsistence between different datasets, and solves this problem by pseudo labels.

\subsection{Generating Auxiliary Data}
Recently with the development of generative models, several works have been proposed and realistic synthetic faces are utilized in training. Some works take advantage of generative models to decouple entangled information such as pose and identity. In \cite{gen-pose1} researchers resolve the variation of face orientation with a generative model, which is able to frontalize a facial image while preserving the identity and expression details.A corresponding framework is proposed in \cite{gen-pose2} where the generative model is used to recognize expressions in different views, conducting hard-sample mining following \cite{deep-adv}. Another twin methods aim at solving the problems of disentangling identity and expression. In \cite{gen-id1} average faces of each emotion are synthesized to eliminate identity differences, while a de-expression process is applied in \cite{gen-id3} to remove interference from subject variations by the proposed residual learning.

Other methods tend to utilize synthetic images to assist training. In \cite{gen-cgan}, researchers use generative models to re-balance the class distribution of a dataset. In \cite{gen-id2}, a facial image will be transformed using Conditional-GAN \cite{cgan} to all emotions contained in the dataset and the model then is trained with those synthetic emotions. The loss function is based on features extracted from both original and generated images, expanding the margin between different classes. Our method is more similar to this fashion, where an intra-class neighborhood transformation will be applied to conduct a dynamic samples selection and reject the unreliable samples, which is shown in the next section.

\section{APPROACH}\label{APP}

\subsection{Problem Formulation}\label{pf}
Let $\mathcal{F}(\cdot)$ represent the prediction model, which is the function mapping an image $I^o$ (in pixel space) to a normalized $K$-dimensional probability vector $P^o$, where $K$ is the number of classes. Suppose we have another image $I^{tr}$, which is almost the same as $I^o$ except some semantic changes, and the corresponding output probability vector is $P^{tr}$. According to our assumption, the divergence of $P^o$ and $P^{tr}$ needs to be relatively small because $I^{tr}$ is exactly a semantic neighbor of $I^o$. The output should be stable to this change, and such a neighborhood transformation of input image shall also lead to small fluctuation in output space, which is noted by:
\begin{equation}
\begin{split}
    S(I^o, I^{tr}) ={}& Div(\mathcal{F}(I^o), \mathcal{F}(I^{tr}))\\
    ={}& Div(P^o, P^{tr}),
\end{split}\label{eqs}
\end{equation}
where $Div(\cdot, \cdot)$ is the metric function to represent the difference of two output probability vector.

$S(I^o, I^{tr})$ substantially indicates neighborhood smoothness of the prediction model, around the given sample $I^o$. A large $S(I^o, I^{tr})$ means a negligible perturbation applied on the semantic space will lead to a significant fluctuation on the output probability vector, which is against the consistency. Therefore, we can distinguish the unreliable samples according to this indicator, leading to a novel two-phase training method, shown in Fig~\ref{overview}.

In the first phase, to synthesize the semantic neighbor $I^{tr}$, we conduct a transformation limited to a neighborhood of the original sample $I^o$.
Tackling high-dimensional data is difficult. Besides, our goal is to generate a semantically-meaningful transformation, which can barely be achieved by perturbation on pixel level, and therefore the difficulty further increases.
To solve the problems of dimension and semantics, we first project the $I^o$ into a latent semantic space, mapping the $I^o$ to a latent representation $z^o$ with typically much smaller dimensionality.
We then apply additive stochastic noise to $z^o$ to get $z^{tr}$, which can be seen as a latent representation of the semantic neighbor. The semantic neighbor can finally be synthesized by reconstructing $I^{tr}$ from $z^{tr}$. Details of formulating the mapping function and reconstructing realistic results from $z^{tr}$ are given in Section~\ref{GN}.


In the second phase, with the semantic neighbor $I^{tr}$, we can then compute $S(I^o, I^{tr})$ for every given $I^o$. As we supposed, images with large $S(\cdot, \cdot)$ should be treated as unreliable samples, and the negative effect of them should be reduced. A threshold can be set to reject the unreliable samples with a larger $S(\cdot, \cdot)$ and gain some performance. However, to set an appropriate threshold needs a tedious process of observation and validation, and the ideal threshold can be change during the training period. We therefore propose a training strategy to estimate the threshold for every training step, described in Section~\ref{DBDS}.

It should be noted that we also take real small facial variation of the same expression into consideration, which can be provided by sequential datasets \cite{oulu}, \cite{ck+}. For a specific frame, other frames near to it are also able to be seen as semantic neighbors. Because only limited kinds of semantic neighbor can be provided by those sequential datasets, we choose to only utilize them in validation to further prove the effectiveness of our method, shown in Section~\ref{EXP_ablation}.

\subsection{Neighborhood Semantic Transformation}\label{GN}

There is no available dataset that can provide corresponding neighbor samples $I^{tr}$ for every $I^o$ because the distributions of samples collected in all datasets are relatively sparse compared with such a high dimensional space. We therefore need to synthesize $I^{tr}$ ourselves. However, images in pixel space are quite high dimensional data therefore hard to tackle and sometime may incur the curse of dimensionality \cite{dimension1}, \cite{dimension2}. And, the perturbation applied on pixel level may not align with the case in semantic space, which means it can not really achieve any types of semantic transformation in facial images, such as raising of lip or eyebrows. We therefore need a mapping function to project the images into a low-dimensional semantic space, as well as a corresponding function to reconstruct the images, keeping realistic.
\begin{table}[tp]
\caption{A detailed description of the architecture of asymmetric autoencoder. (Shape of output of convolutional layer is described as [channel, height, width])}
\label{arch_ae}
\begin{center}
    \begin{tabular}{|>{\centering}p{41pt}|>{\centering}p{19pt}|c|>{\centering}p{17pt}|c|}
    \hline
    \textbf{Layer}&\textbf{Kernel}&\textbf{Output}&\textbf{Stride}&\textbf{Other Setting}\\
    \hline
    PreProcess&-&[1, 112, 112]&-&DP$^\mathrm{1}$\\
    \hline
    E-Conv-a&$3\times3$&[16, 56, 56]&2&BN$^\mathrm{2}$\&LReLu$^\mathrm{3}$\&DP\\
    \hline
    E-Conv-b&$3\times3$&[32, 28, 28]&2&BN\&LReLu\&DP\\
    \hline
    E-Conv-c&$3\times3$&[64, 14, 14]&2&BN\&LReLu\&DP\\
    \hline
    E-Conv-d&$3\times3$&[80, 7, 7]&2&BN\&Tanh\\
    \hline
    E-FC&-&64&-&-\\
    \hline
    D-FC&-&$80\times7\times7$&-&-\\
    \hline
    D-TConv-c$^\mathrm{4}$&$3\times3$&[64, 14, 14]&2&$2\times$Res$^{\mathrm{5}}$\&ReLu\\
    \hline
    D-TConv-b&$3\times3$&[32, 28, 28]&2&$2\times$Res\&ReLu\\
    \hline
    D-TConv-a&$3\times3$&[16, 56, 56]&2&$2\times$Res\&ReLu\\
    \hline
    D-TConv-t&$3\times3$&[16, 112, 112]&2&$2\times$Res\&ReLu\\
    \hline
    D-TConv-o&$3\times3$&[1, 112, 112]&2&$1\times$Res\\
    \hline
    \multicolumn{5}{l}{$^{\mathrm{1}}$Dropout with probability of 0.1. $^{\mathrm{2}}$Batch Normalization.}\\
    \multicolumn{5}{l}{$^{\mathrm{3}}$LeakyReLu, $\alpha=0.01$. $^{\mathrm{4}}$Transpose Conv. $^{\mathrm{5}}$Stacked ResBlock\cite{res}.}
    \end{tabular}
\end{center}
\end{table}

\noindent{Projecting to Low-dimensional Semantic Space:}
Autoencoder has been proven to be effective in producing semantically meaningful and well-separated representations on real-world datasets by plenty of research \cite{ae1}, \cite{ae2}, \cite{ae3}. Therefore, we propose to train a denoising stacked autoencoder and thereby learn a semantic low-dimensional representation using the encoder.

\begin{figure*}[tp]
\centering
\begin{minipage}{0.07\linewidth}
\includegraphics[width=\linewidth]{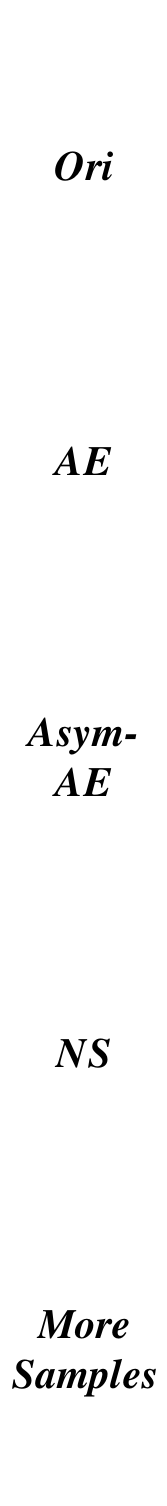}
\centerline{ }
\end{minipage}
\begin{minipage}{0.124\linewidth}
\animategraphics[loop, width=\textwidth]{6}{img/aff/figure_aff_angry-}{0}{5}
\centerline{Anger}
\end{minipage}
\begin{minipage}{0.124\linewidth}
\animategraphics[loop, width=\textwidth]{6}{img/aff/figure_aff_disgust-}{0}{5}
\centerline{Disgust}
\end{minipage}
\begin{minipage}{0.124\linewidth}
\animategraphics[loop, width=\textwidth]{6}{img/aff/figure_aff_fear-}{0}{5}
\centerline{Fear}
\end{minipage}
\begin{minipage}{0.124\linewidth}
\animategraphics[loop, width=\textwidth]{6}{img/aff/figure_aff_happy-}{0}{5}
\centerline{Happiness}
\end{minipage}
\begin{minipage}{0.124\linewidth}
\animategraphics[loop, width=\textwidth,]{6}{img/aff/figure_aff_neutral-}{0}{5}
\centerline{Neutral}
\end{minipage}
\begin{minipage}{0.124\linewidth}
\animategraphics[loop, width=\textwidth]{6}{img/aff/figure_aff_sad-}{0}{5}
\centerline{Sadness}
\end{minipage}
\begin{minipage}{0.124\linewidth}
\animategraphics[loop, width=\textwidth]{6}{img/aff/figure_aff_surprise-}{0}{5}
\centerline{Surprise}
\end{minipage}
\caption{
Generated results on AffectNet database. Neighborhood transformation can be perceived on emotion-relative area, e.g. lips and eyebrows, while those irrelevant regions are filled mistily. \emph{Ori}: original inputs grayed and resized to $112\times112$ pixels. \emph{AE}: results of autoencoder with MSE reconstruction loss only. \emph{Asym-AE}: final generated results with asymmetric autoencoder. \emph{NS}: a neighbor sample synthesized with stochastic additive noise. \emph{More Samples}: more synthesized neighbor samples. \textbf{The perturbation is difficult to be observed, and we therefore make it play as video on click. Best viewed in Adobe Reader.}}
\label{aff_show}
\end{figure*}

The encoder $\mathcal{H}_e$ maps noisy data into an embedding subspace with several convolutional layers followed by a fully connected layer. The output of \emph{l}-th convolutional layer can be indicated by the following equation:
\begin{equation}
    \Tilde{z}^l = Dropout[g(W^l_e\Tilde{z}^{l-1})],
\end{equation}
where $Dropout(\cdot)$ is the stochastic mask function and $W^l_e$ indicates the weights of the \emph{l}-th layers. $\Tilde{z}^l$ are the features of \emph{l}-th layer and $\Tilde{z}^0$ are equal to the noisy input data $\Tilde{x}$. $g(\cdot)$ is activation function, which is in fact the Leaky-ReLU function here.

Following the corrupted encoder, a decoder $\mathcal{H}_d$ is built to reconstruct the input data from the output features of the fully connected layer at first. Conventionally, the structure of decoder is symmetric with that of encoder, recovering outputs of each layers in encoder with a fully connected layer and several transpose convolutional layers respectively:
\begin{equation}
    \Hat{z}^{l-1} = g(W^l_d\Hat{z}^{l}).
\end{equation}
$\Hat{z}^{l}$ here denotes the recovered results of $\Tilde{z}^l$ and $W^l_d$ shows the weights of corresponding layer in decoder. The reconstruction of input $\Hat{x}$ is $\Hat{z}^0$ indeed.

We do not train the autoencoder in a layer-wise fashion but jointly train all layers together following \cite{cluster1}, and the loss function is defined as:
\begin{equation}
    Loss_{AE} =  \sum\limits_{l=0}^{L-1} \frac{\alpha_l}{|z^l|}||z^l-\Hat{z}^l||^2_2 \label{l_ae},
\end{equation}
where $|z^l|$ is the feature size of the \emph{l}-th scale level, $\alpha_l$ is the weighting coefficient of recovering $z^l$ and $L$ is the depth of encoder.

\noindent{\textbf{{Reconstructing Realistic Semantic Neighbor:}}
With the usage of autoencoder, we preliminarily solved the two problems of neighborhood semantic transformation. However, images reconstructed by autoencoder cannot be utilized as $I^{tr}$ directly and need further refinement. 
Those images are fuzzy in usual, as shown in the second row of Fig~\ref{aff_show}, and we propose that the problems to solve are three-fold:
\begin{enumerate}
\item The recovering performance of decoder is unsatisfactory and requires further improvement.
\item Squared Euclidean (SE) distance we used for reconstruction loss can lead to blurred results and need to be replaced.
\item Another metric needs to be formulated to
synthesize more realistic samples.
\end{enumerate}

We remedy the three problems respectively as follows. For the first requirement, to enhance the reconstruction performance of autoencoder we thicken the decoder by replacing all convolutional layers with ResBlock\cite{res} and add an extra one after each block. Two extra blocks are added before the output block for better refinement. Those adaptations lead to an asymmetric autoencoder, and the final architecture is shown in TABLE~\ref{arch_ae}.

For the second, to keep the consistency of $I$ and $\mathcal{H}_d(\mathcal{H}_e(I))$ as well as reduce the blurring, we use the weighted sums of perceptual metric and SE metric \cite{perceptual} as a substitute to form the final reconstruction loss:
\begin{equation}
\begin{split}
    Loss_{Rec} = {}&\lambda_{pixel}||\mathcal{H}_d(\mathcal{H}_e(I))-I||_2^2\\
            {}&+\lambda_{perc}||\mathcal{C}(\mathcal{H}_d(\mathcal{H}_e(I)))-\mathcal{C}(I)||_2^2,
\end{split}
\end{equation}
where $\mathcal{C}(\cdot)$ denotes a well-trained convolutional network.

Apart from that, for the last problem, we finally use adversarial loss as a tool to attain a more realistic and sharp result, which is in fact the value function advocated by WGAN\cite{wgan}. The corresponding loss function used in discriminator $\mathcal{D}(\cdot)$ and generator(the decoder $\mathcal{H}_d(\cdot)$) are defined as:
\begin{equation}
    Loss^D_{Adv} = \mathcal{D}(\mathcal{H}_d(\mathcal{H}_e(I^u))-\mathcal{D}(I^v),
\end{equation}
\begin{equation}
    Loss^G_{Adv} = -\mathcal{D}(\mathcal{H}_d(\mathcal{H}_e(I^u))),
\end{equation}
where $I^u$ and $I^v$ both belong to the real-world dataset.

We train the asymmetric autoencoder first using $Loss_{AE}$ to get parameters sufficiently pre-trained. Then we use all loss functions jointly and train the discriminator as well. The final loss function of the discriminator is exactly $Loss^D_{Adv}$, and that of decoder is defined as:
\begin{equation}
\begin{split}
    Loss_G ={}&\lambda_{AE}Loss_{AE}+\lambda_{Adv}Loss^G_{Adv}\\
    {}&+\lambda_{Rec} Loss_{Rec}.
\end{split}
\end{equation}

Following this process, we finally gain the $\mathcal{H}_d$ and $\mathcal{H}_e$ that can both embed facial images into low-dimensional semantic space and reconstruct them from the embedding codes. Thereby, for any given $I^o$, the semantic neighbor $I^{tr}$ can be formulated as:
\begin{equation}
\label{gen_eq}
    I^{tr} = \mathcal{H}_d(\mathcal{H}_e(I^o)+\mathcal{N}),
\end{equation}
where $\mathcal{N}$ denote a stochastic additive noise, and we simply use a Gaussian noise in this paper. We then can rewrite \eqref{eqs}, the indicator of unreliable samples as:
\begin{equation}
\label{s()}
    S(I^o) = Div(\mathcal{F}(I^o), \mathcal{F}(\mathcal{H}_d(\mathcal{H}_e(I^o)+\mathcal{N}))).
\end{equation}
Fig~\ref{overview}-(a) illustrates the process of synthesizing a semantic neighbor using our asymmetric autoencoder. Training details can be seen in Section \ref{EXP} and the synthesized results are shown in Fig~\ref{aff_show}.

\subsection{Semantic Neighborhood-Aware Optimization} \label{DBDS}
With our semantic neighborhood transformation, given a sample $I^o$ we can compute $S(I^o)$ by \eqref{s()} naturally. In this section we focus on how to utilize the $S(I^o)$ to guide the optimization, or more specially, how to distinguish the unreliable samples using $S(I^o)$ and what to do with them.

\noindent{\textbf{{Threshold-based Optimization:}}
$S(\cdot)$ can be seen as an indicator of unreliable samples. During the training period, the negative effect of them should be reduced to gain better consistency of input and output change, as well as performance. Therefore, we propose an optimization method with threshold at first, simply rejecting the samples with $S(\cdot)$ larger than a preset threshold.

However, $S(I^o)$ is not an immutable property of the given $I^o$ because it changes ceaselessly along with the parameters of prediction model $\mathcal{F}$ when training. Therefore it is inappropriate to make the rejected $I^o$ be unseen by the model in all subsequent training and the estimation of unreliable samples should be repeated in every training step.

To formulate our method, given a sample $I^o$, we compute $S(I^o)$ using the current parameters of prediction model $\mathcal{F}$ following \eqref{s()} firstly. After that, we simply mask the gradient computed from those samples with $S(\cdot)$ larger than the threshold $T$, and hence these samples contribute no stimulation to the prediction model in this training step.
Let $P^{\{\cdot\}}$ be the output probability vector, then the loss of $Loss_{T}$ is defined as:
\begin{equation}
\label{T_dtds_eq}
    Loss_{T} =
    -\delta(T-Div(P^o, P^{tr}))\sum_{i=0}^K y_i ln(P^o_i),
\end{equation}
where $y$ is the ground-truth label and the subscript $i$ indicates different classes. $\delta(\cdot)$ is the indicator function where $\delta(x)=1$ if $x > 0$ else $\delta(x)=0$. We use a symmetric variant of the Kullback--Leibler divergence to formulate the metric $Div(\cdot, \cdot)$ in this paper:
\begin{equation}
    KL(P||Q) = \sum_i P_i log(\frac{P_i}{Q_i}),
\end{equation}
\begin{equation}
\label{eq4}
    Div(P^o, P^{tr}) = \frac{1}{2}(KL(P^o||P^{tr})+KL(P^{tr}||P^o)).
\end{equation}

The training procedure are illustrated in the middle of Fig~\ref{overview} without the dashed arrow. More details can be seen in Section \ref{EXP}.

\noindent{\textbf{{Batch-level Threshold-based Optimization:}}
A simply threshold alleviates the problem in a way but still leaves two flaws to be remedy. The threshold $T$ is estimated experimentally, which needs a tedious process of observation and experiment. With a large $T$, all samples will be accepted and therefore our algorithms fails, while with a small $T$ most of the samples will be rejected and the network then cannot be trained sufficiently. What is worse, the appropriate threshold can change during the training, thus a constant threshold is not suitable for all the periods.

We proposed a mask on batch level to fix those problems by computing a threshold in each training mini-batch and applying it to the samples in those mini-batches respectively. We here use the mean of a mini-batch to do so:
\begin{equation}
\label{T_dbds_eq}
    T_{batch} = \frac{1}{N_{batch}}\sum_{P^o \in batch}Div(P^o, P^{tr}),
\end{equation}
where $N_{batch}$ is the size of training mini-batch. The pseudo-code of our method is shown in Algorithm~\ref{alg_DBDS}. Section~\ref{ablation} shows detail performance comparison of these two method.
\begin{algorithm}[htp]
	\renewcommand{\algorithmicrequire}{\textbf{Input:}}
	\renewcommand{\algorithmicensure}{\textbf{Output:}}
	\caption{Training Algorithm of Our Method}
	\label{alg_DBDS}
	\begin{algorithmic}[1]
		\REQUIRE Training data $\{I^i, y^i\}^n_i$, where $n$ is the size of mini batch; Well-trained asymmetric autoencoder $\mathcal{H}_d, \mathcal{H}_e$; Learning rate $\mu$; Flag $is\_batch$; Threshold $T$;
		\ENSURE Parameters $W$
		\STATE Initialize $t \leftarrow 0$
		\WHILE{$t<t_{max}$}
		\FORALL{$I^i$}
		\STATE Generate transformed image $I^{i;tr}$ by \eqref{gen_eq}
		\STATE Compute probability vector $P^i$ of image $I^i$ with parameters $W^t$
		\STATE Compute cross entropy loss $L^i$
		\STATE Compute probability vector $P^{i;tr}$ of image $I^{i;tr}$ with parameters $W^t$
		\STATE Compute divergence $Div(P^i, P^{i;tr})$ by \eqref{eq4}
		\ENDFOR
		\IF{$is\_batch$}
		\STATE Compute threshold $T$ for current mini-batch by \eqref{T_dbds_eq}
		\ENDIF
		\STATE Compute mask coefficient $m^i$ for each $i$:\\
		$m^i \leftarrow \delta(T-Div(P^i, P^{i;tr}))$
		\STATE Update parameters $W^{t+1}$:\\
		$W^{t+1} \leftarrow W^t - \mu^t\sum_{i=0}^n m^i\frac{\partial L^i}{\partial W^t}$
		\STATE $t \leftarrow t+1$
		\ENDWHILE
		\STATE \textbf{return} $W^{t_{max}}$
	\end{algorithmic}
\end{algorithm}
\begin{table*}[tp]
\caption{A detailed description of the architecture of proposed multiscale network. (Shape of output of convolutional layer is described as [channel, height, width])}
\label{arch_pred}
\begin{center}
    \begin{tabular}{|c|c|c|c|c|c|c|c|}
    \hline
    \textbf{Layer}&\textbf{Kernel}&\textbf{Output}&\textbf{Layer}&\textbf{Kernel}&\textbf{Output}&\textbf{Pooling}&\textbf{Other Setting}\\
    \hline
    PreProcess-Ori&-&[1, 112, 112]&PreProcess-Half&-&[1, 56, 56]& $2\times2$&-\\
    \hline
    Ori-Conv1-a&$5\times5$&[64, 56, 56]&Half-Conv1-a&$5\times5$&[64, 28, 28]&$2\times2$&BN\&ReLu\\
    Ori-Conv1-b&$5\times5$&[64, 56, 56]&Half-Conv1-b&$5\times5$&[64, 28, 28]&$2\times2$&BN\&ReLu\\
    \hline
    Ori-Conv2-a&$3\times3$&[128, 28, 28]&Half-Conv2-a&$3\times3$&[128, 14, 14]&$2\times2$&BN\&ReLu\\
    Ori-Conv2-b&$3\times3$&[128, 28, 28]&Half-Conv2-b&$3\times3$&[128, 14, 14]&$2\times2$&BN\&ReLu\\
    \hline
    Ori-Conv3-a&$3\times3$&[256, 14, 14]&Half-Conv3-a&$3\times3$&[256, 7, 7]&$2\times2$&BN\&ReLu\\
    Ori-Conv3-b&$3\times3$&[256, 14, 14]&Half-Conv3-b&$3\times3$&[256, 7, 7]&$2\times2$&BN\&ReLu\\
    \hline
    Ori-Conv4&$3\times3$&[512, 7, 7]&Half-Conv4&$3\times3$&[512, 3, 3]&$2\times2$&BN\&ReLu\\
    \hline
    Ori-Conv5&$1\times1$&[128, 7, 7]&Half-Conv5&$1\times1$&[128, 3, 3]&$2\times2$&BN\&ReLu\&DP$^{\mathrm{1}}$\\
    \hline
    Ori-FC&-&64&Half-FC&-&32&-&-\\
    \hline
    Fuse-FC&-&7&-&-&-&-&-\\
    \hline
    \multicolumn{5}{l}{$^{\mathrm{1}}$Dropout with probability of 0.5.}
    \end{tabular}
\end{center}
\end{table*}

\section{EXPERIMENTS}\label{EXP}
We validate the effectiveness of our algorithm on CK+\cite{ck+}, Oulu-CASIA\cite{oulu}, and AffectNet\cite{affectnet}, all of which are widely used. Specifically, we mainly use AffectNet for further analysis and discussion because it is now the largest dataset with annotated in-the-wild facial emotions, which makes it more stable and persuasive to validate the performance.

A description of those three datasets is given in Section~\ref{EXP_data} first, and implementation details about network architecture, hyper-parameters and other tricks are shown in Section~\ref{EXP_detail} to provide a guarantee for reproduction. Results of both generation model and prediction model compared with state-of-the-art methods are shown in Section~\ref{EXP_trans} and Section~\ref{EXP_soa}, respectively. Analysis
of the control experiment and ablation study can be found in Section~\ref{EXP_ablation}.
\subsection{Datasets Setting}\label{EXP_data}

\noindent{\textbf{{Extended CohnKanade:}}
The Extended CohnKanade (CK+) dataset, which is laboratory-controlled, is the most extensively used. CK+ contains 593 video sequences recorded from 123 subjects, where 118 subjects with 327 sequences are labelled. Following the widely used evaluation protocol in \cite{dcn+ap}, we select the last three frames of each sequence labeled with basic emotions (\emph{Anger}, \emph{Disgust}, \emph{Fear}, \emph{Happiness}, \emph{Sadness}, \emph{Surprise}) for both training and testing. Landmarks of each frame are provided, thus we can crop frames using the coordinates of them with a margin of $15\%$.

\noindent{\textbf{{Oulu-CASIA:}}
Oulu-CASIA is another commonly used in-the-lab dataset for FER, containing videos recorded from 80 subjects. Each subject is captured with six basic emotions by both NIR and VIS cameras respectively.
For a better comparison, we use the cropped subset provided in \cite{oulu} and select the last three frames of each sequence for validation. Apart from that, according to \cite{principle}, all the experiments we conduct in this paper are following a person-independent fashion.

\noindent{\textbf{{AffectNet:}}
AffectNet is the largest in-the-wild FER dataset with more than 400,000 images collected from three search engines, annotated manually as 10 categories (six basic emotion plus \emph{Neutral}, \emph{Contempt}, \emph{None}, \emph{Uncertain} and \emph{Non-face}). A validation set is supported with 500 samples for each class. For a fair comparison, we use the images with \emph{Neutral} and other six basic emotion labels, which lead to around 280,000 samples in 400,000 and 3500 samples in 5000.

As mentioned above, all these three datasets have their own problems to be solved, hindering the performance of training models. For CK+ and Oulu-CASIA, the main impediment is the lack of data, for there are only around 300 subjects to train a model and 30 subjects to validate in CK+ in each 10-fold validation, and in Oulu-CASIA the numbers change to 432 and 48. The insufficiency of data not only increases the risk of overlearning as well as hinders the performance, but also brings the instability in validation, where only one or two samples can lead to 1\% or even more performance change.

That is why we mainly use AffectNet to validate the performance, though it also has its own problem, i.e. the extreme imbalance in their class distribution. The majority classes, e.g. \emph{Happiness} and \emph{Neutral}, make up 73.7\% of the whole dataset while the minority classes such as \emph{Fear} and \emph{Disgust} only occupy a proportion of 3.6\% together. Apart from that, each image in the training subset is labeled by only one annotator, and annotators may take issue with each other on labeling data. According to \cite{affectnet}, annotators only agree on 60.7\% of images for all 10 categories and 65.4\% for the 7 classes we used. Details are listed in \cite{affectnet} and this rate of agreement might be the ceiling of prediction performance in AffectNet.
\begin{table}[t]
\caption{Expression recognition average accuracy of different methods on Affectnet database. (\textbf{Bold}: Best result. \underline{Underline}: Second best.)}\label{aff_score}
\begin{center}
    \begin{tabular}{|c|c|}
    \hline
    \textbf{METHOD}&\textbf{ AVERAGE ACCURACY}\\
    \hline
    Annotators Agreement\cite{affectnet}&65.3\%\\
    \hline
    IPA2LT\cite{trans-inconsis}&57.31\%\\
    VGG16\cite{vgg}&51.11\%\\
    DLP-CNN\cite{dlp}&54.47\%\\
    pACNN\cite{prior-landmark}&55.33\%\\
    gACNN\cite{prior-landmark}&58.78\%\\
    FMPN\cite{fmpn}&\textbf{61.5\%}\\
    EAU-NET\cite{gen-cgan}&58.91\%\\
    PG-CNN\cite{pgcnn}&55.33\%\\
    VGG-FACE\cite{vgg-face}&\underline{60.0\%}\\
    \hline
    \textbf{Our Method}&\textbf{62.7\%}\\
    \hline
    \end{tabular}
\end{center}
\end{table}
\subsection{Implementation Details}\label{EXP_detail}
\noindent{\textbf{{Details of Neighborhood Semantic Transformation:}}
The process of training can be divided into two stages, i.e. pre-training and refining. Firstly, we pre-train the asymmetric autoencoder with the loss $L_{AE}$ defined in \eqref{l_ae}, which will gain us a good initialization of parameters and blurred recovering results. After that, to generate realistic synthesis images, we use $Loss_G$ and $Loss_{Rec}$ in the second stage for refinement.

For reducing the scale of memory usage, images in all three datasets are scaled to a size of $112\times112$ pixels and are preprocessed by graying while most of other methods use color images with size of $224 \times 224$. The model in the first stage is trained with an SGD optimizer for 50 epochs with momentum of 0.9, weight decay of 0.0004 and mini-batch size of 256. Training learning rate is initialized as $10^{-3}$ and decreases by 0.1 every 10 epochs. As shown in TABLE~\ref{arch_ae}, we set the length of latent code to 64 and have four hidden layers in encoder and nine ResBlocks\cite{res} in decoder, while the number of different scales is 5 in both, including the original input. Weighting coefficients $\alpha_l$ in \eqref{l_ae} are set to 4 for the input images and 1 for features from other hidden layers.

In the refinement stage, when training with $Loss^G_{Adv}$, $Loss_{Rec}$ and $Loss_{AE}$ jointly, we expand the training epochs number to 100 with the mini-batch size decreasing to 64 and the optimizer is replaced with RMSprop where both momentum and weight decay are set to 0 as \cite{wgan} recommended. Learning rates of decoder $\mathcal{H}_d$ and discriminator are fixed to $10^{-4}$ and $2\times10^{-5}$ respectively. We extract the features from $Conv3$ to $Conv5$ in an FER network, which is our well-trained baseline in fact, to compute the perceptual loss with weighting coefficients of $\{100, 0.1, 0.001\}$ respectively. $\lambda_{AE}$ is set to $10^{-3}$ at first and decreases by 0.1 every 20 epochs, while $\lambda_{Adv}$ and $\lambda_{Rec}$ are fixed to 5 and 1 respectively. We train the discriminator once every 10 times the generator get trained except the first epoch.


\begin{table}[tp]
\caption{Expression recognition average accuracy of different methods on Extended Cohnkanade (CK+) database. (\textbf{Bold}: Best result. \underline{Underline}: Second best.)}
\label{ckp_score}
\begin{center}
    \begin{tabular}{|c|c|}
    \hline
    \textbf{METHOD}&\textbf{ AVERAGE ACCURACY}\\
    \hline
    CSPL\cite{cspl}&89.9\%\\
    LBPSVM\cite{lbp} &95.1\%\\
    3DCNN-DAP\cite{dap}&92.4\%\\
    BDBN\cite{bdbn}&96.7\%\\
    PPDN\cite{trans-peak}&97.3\%\\
    Zero-bias CNN\cite{zero}&98.3\%\\
    STM-ExpLet\cite{stm}&94.2\%\\
    Dis-ExpLet\cite{mid-feat1}&95.1\%\\
    DTAGN\cite{prior-landmark2}&97.3\%\\
    LOMO\cite{lomo}&95.1\%\\
    DLP-CNN\cite{dlp}&95.78\%\\
    Inception\cite{cnn3}&93.2\%\\
    FN2EN\cite{trans-f2e}&\underline{98.6\%}\\
    DCN+AP\cite{dcn+ap}&\textbf{98.9\%}\\
    ESL\cite{esl}&95.33\%\\
    \hline
    \textbf{Our Method}&\textbf{98.58\%}\\
    \hline
    \end{tabular}
\end{center}
\end{table}

\noindent{\textbf{{Training Details of FER Model: }}The architecture of our prediction model is shown in TABLE~\ref{arch_pred}, which is a five-stage network with 2 branches of different scales. Both of these two branches can produce a probability vector and the output of last fully connected layers of them will be fused to compute the final prediction result. We train the model using the sum of losses computed by these three prediction with the same weight.
We conduct Selective Learning (SL)\cite{select}, ensuring that every class will occupy equal proportion in each mini-batch to rebalance the classes distribution. Focal loss\cite{focal} is also used to focus training on those misclassified samples.


We train our network using SGD optimizer with a weight decay of 0.002 and a momentum of 0.9 for 50 epochs. The learning rate decreases by 0.1 every 10 epochs starting from an initial value of $10^{-2}$. Batch size is set to 63 for it needs to be a multiple of classes number in SL\cite{select}. Besides, the momentum of Batch Normalization is also set to 0.9. All these settings make our model be adept at tackling this tricky dataset, leading to a powerful baseline, which is comparable to the state of the art.

\subsection{Results of Transformation}\label{EXP_trans}

\begin{figure*}[t]
\centering
\begin{minipage}{0.07\linewidth}
\includegraphics[width=\linewidth]{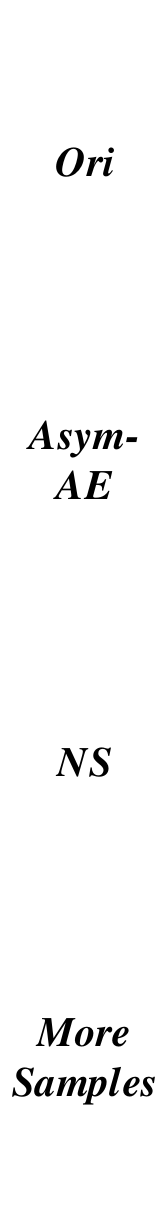}
\centerline{ }
\end{minipage}
\begin{minipage}{0.12\linewidth}
\animategraphics[loop, width=\textwidth]{6}{img/ckp/figure_ck_angry-}{0}{5}
\centerline{Anger}
\end{minipage}
\begin{minipage}{0.12\linewidth}
\animategraphics[loop, width=\textwidth]{6}{img/ckp/figure_ck_disgust-}{0}{5}
\centerline{Disgust}
\end{minipage}
\begin{minipage}{0.12\linewidth}
\animategraphics[loop, width=\textwidth]{6}{img/ckp/figure_ck_fear-}{0}{5}
\centerline{Fear}
\end{minipage}
\begin{minipage}{0.12\linewidth}
\animategraphics[loop, width=\textwidth]{6}{img/ckp/figure_ck_happy-}{0}{5}
\centerline{Happiness}
\end{minipage}
\begin{minipage}{0.12\linewidth}
\animategraphics[loop, width=\textwidth]{6}{img/ckp/figure_ck_sad-}{0}{5}
\centerline{Sadness}
\end{minipage}
\begin{minipage}{0.12\linewidth}
\animategraphics[loop, width=\textwidth]{6}{img/ckp/figure_ck_surprise-}{0}{5}
\centerline{Surprise}
\end{minipage}
\caption{
Generated results on Extended CohnKanade (CK+) database. Neighborhood transformation can be perceived on emotion-relative area such as lips and eyebrows, while those irrelevant regions are filled mistily. \emph{Ori}: original inputs grayed and resized to $112\times112$ pixels. \emph{Asym-AE}: final generated results with asymmetric autoencoder. \emph{NS}: a neighbor sample synthesized with stochastic additive noise. \emph{More Samples}: more synthesized neighbor samples. \textbf{The perturbation is difficult to be observed, and we therefore make it play as video on click. Best viewed in Adobe Reader.}}
\label{ckp_show}
\end{figure*}

\begin{figure*}[t]
\centering
\begin{minipage}{0.07\linewidth}
\includegraphics[width=\linewidth]{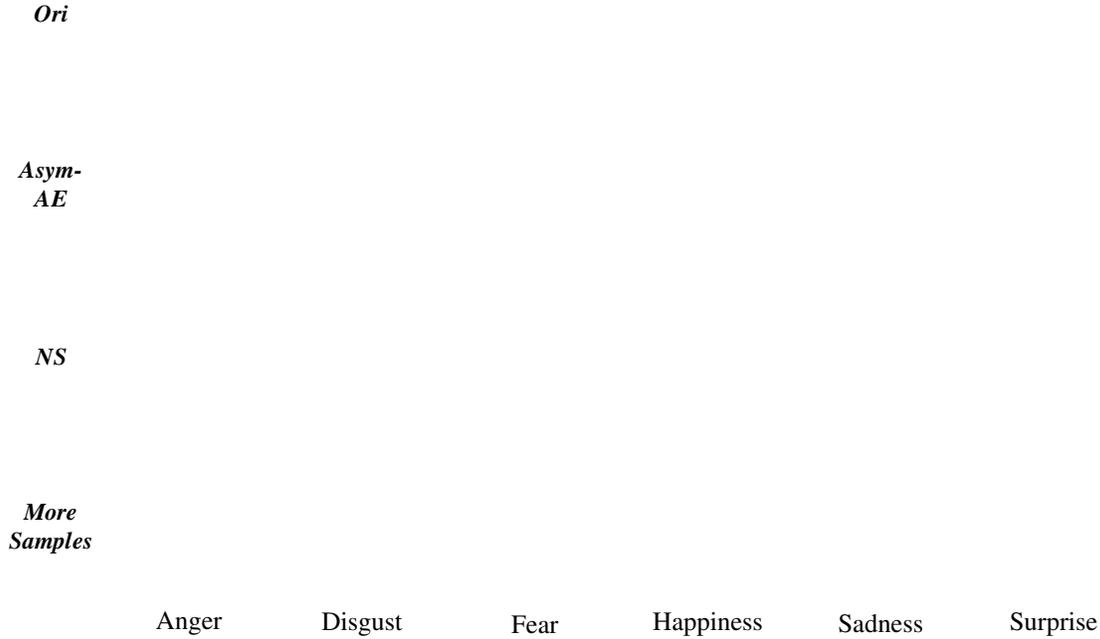}
\centerline{ }
\end{minipage}
\begin{minipage}{0.12\linewidth}
\animategraphics[loop, width=\textwidth]{6}{img/oulu/figure_oulu_angry-}{0}{5}
\centerline{Anger}
\end{minipage}
\begin{minipage}{0.12\linewidth}
\animategraphics[loop, width=\textwidth]{6}{img/oulu/figure_oulu_disgust-}{0}{5}
\centerline{Disgust}
\end{minipage}
\begin{minipage}{0.12\linewidth}
\animategraphics[loop, width=\textwidth]{6}{img/oulu/figure_oulu_fear-}{0}{5}
\centerline{Fear}
\end{minipage}
\begin{minipage}{0.12\linewidth}
\animategraphics[loop, width=\textwidth]{6}{img/oulu/figure_oulu_happy-}{0}{5}
\centerline{Happiness}
\end{minipage}
\begin{minipage}{0.12\linewidth}
\animategraphics[loop, width=\textwidth]{6}{img/oulu/figure_oulu_sad-}{0}{5}
\centerline{Sadness}
\end{minipage}
\begin{minipage}{0.12\linewidth}
\animategraphics[loop, width=\textwidth]{6}{img/oulu/figure_oulu_surprise-}{0}{5}
\centerline{Surprise}
\end{minipage}
\caption{Generated results on Oulu-CASIA database. Neighborhood transformation can be perceived on emotion-relative area such lips and eyebrows, while those irrelevant regions are filled mistily. \emph{Ori}: original inputs grayed and resized to $112\times112$ pixels. \emph{Asym-AE}: final generated results with asymmetric autoencoder. \emph{NS}: a neighbor sample synthesized with stochastic additive noise. \emph{More Samples}: more synthesized neighbor samples. \textbf{The perturbation is difficult to be observed, and we therefore make it play as video on click. Best viewed in Adobe Reader.}}
\label{oulu_show}
\end{figure*}

Fig~\ref{aff_show} illustrates several samples of the generated face images of six basic emotions plus \emph{Neutral} on AffectNet. The first line contains the original input images, which are all grayed and resized to a size of $112\times112$ pixels. The second line is the result of autoencoder, without the following training process with adversarial loss, leading to pretty fuzzy output. Our final generated results with/without additive noise are shown in the third and the fourth lines respectively. Images in fifth line are made play as videos thus the perturbation can be observed more easily. \emph{All videos play on click.}

It can be seen that background area and even some regions which have nothing to do with expression recognition, e.g. hairs and clothes, are all filled with average color and texture. That is because the model we used for computing perceptual loss is a well-trained FER network. Therefore, the change of pixels on the facial area will evoke a fiercer response on feature maps of the FER network, while distortion on other regions will be almost ignored compared with the former. Though the performance of verisimilitude may get harmed, it benefits our goal because more information associated with facial emotions can be well preserved by hidden representations. As shown in the last line of Fig~\ref{aff_show}, additive noise applied to the hidden representations does incur neighborhood transformation in the synthesized images, varying from illumination variations to the change of lips, eyebrows and facial contour.

Fig~\ref{oulu_show} and Fig~\ref{ckp_show} illustrate samples of synthesized facial images of six basic emotions on Oulu-CASIA and CK+, respectively. The last line can also play on click. As the same with results of AffectNet, neighborhood transformation can be perceived on emotion-relative area such as lips and eyebrows, while those irrelevant regions are filled mistily.

\subsection{Comparison with Other FER Methods}\label{EXP_soa}

\begin{figure}[htp]
\centering
\begin{minipage}{0.49\linewidth}
\includegraphics[width=\linewidth]{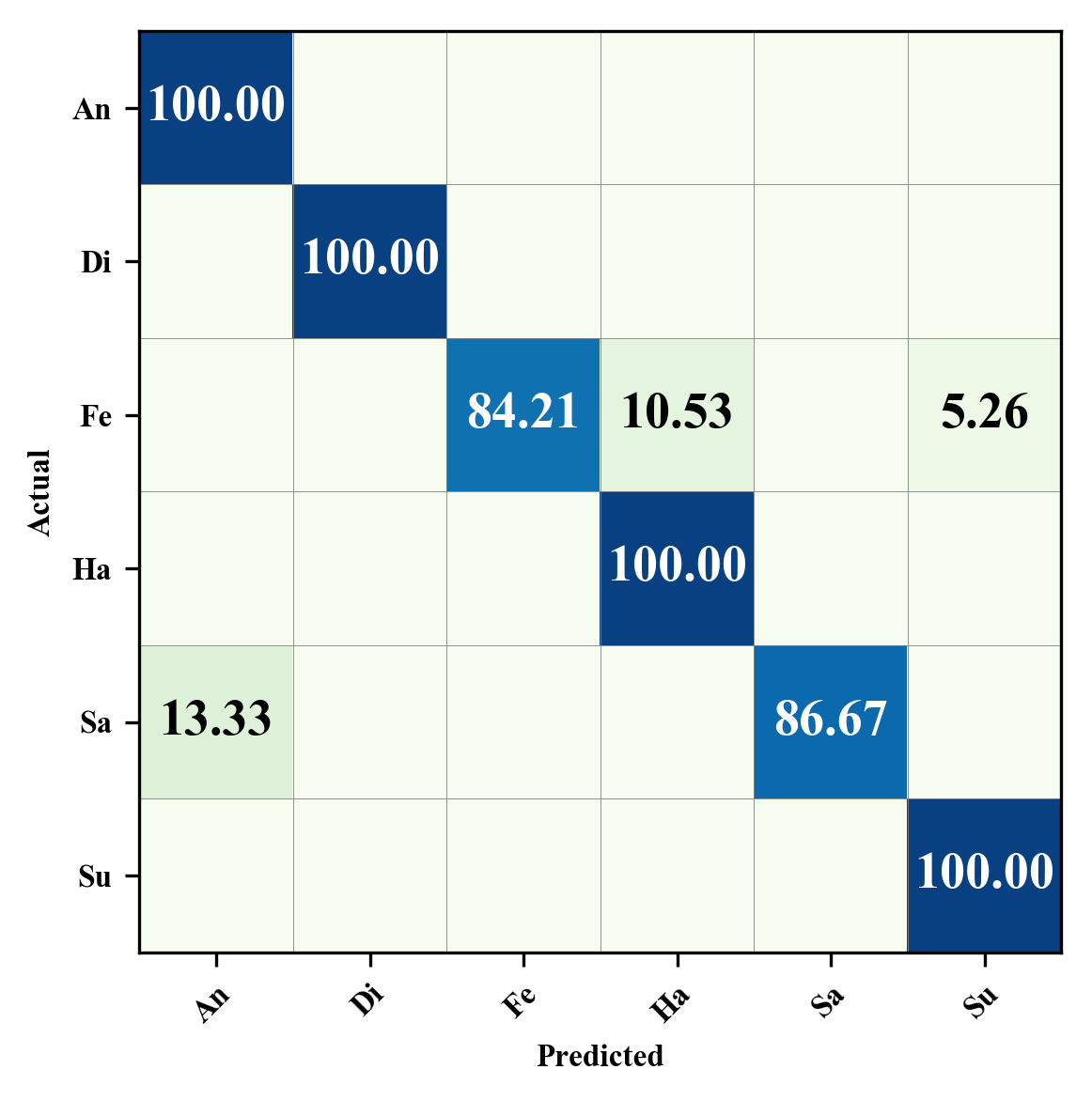}
\centerline{(a) CK+}
\end{minipage}
\begin{minipage}{0.49\linewidth}
\includegraphics[width=\linewidth]{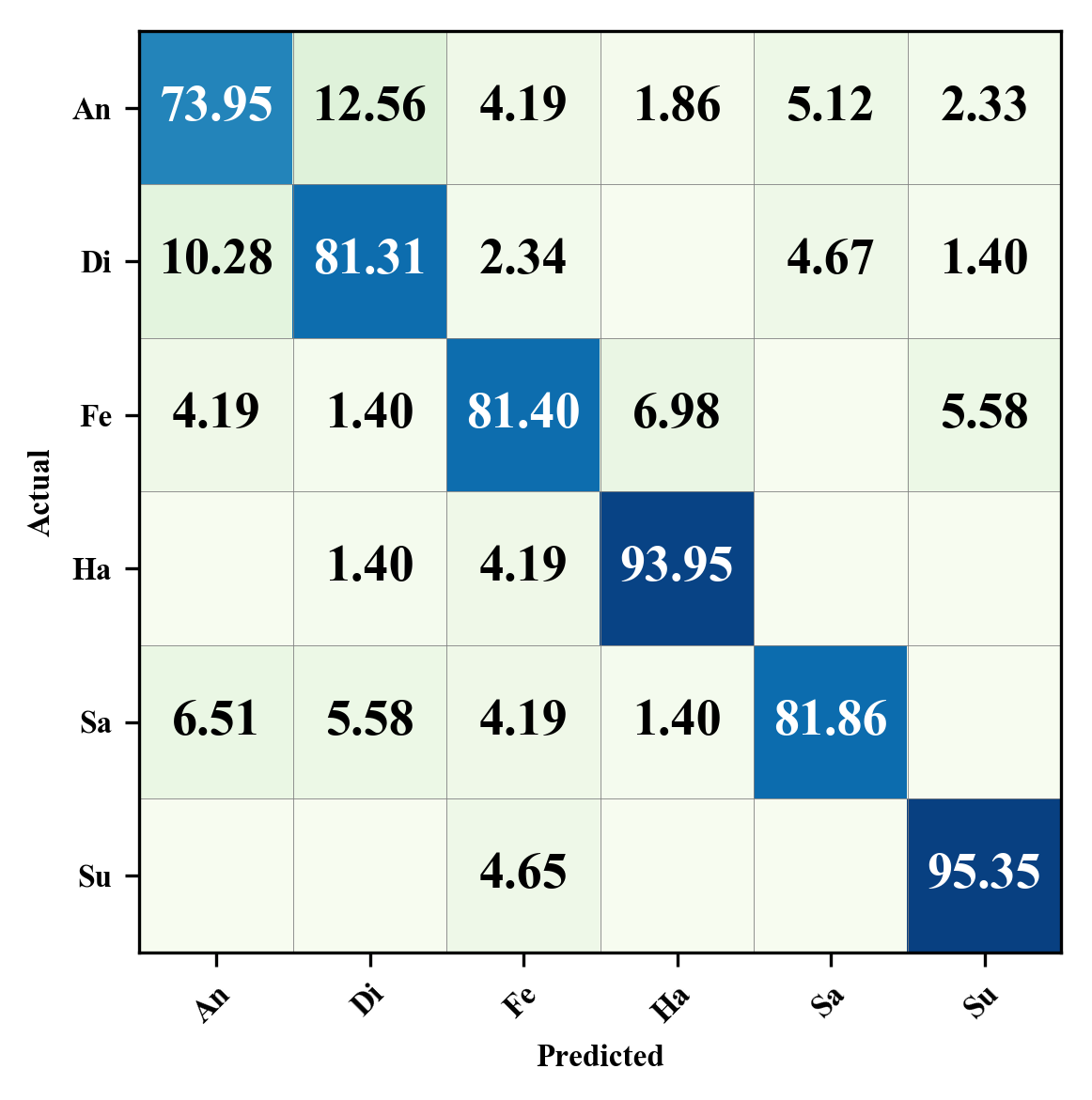}
\centerline{(b) Oulu-CASIA}
\end{minipage}
\caption{
(a): confusion matrix of our multi-scale model with our method on CK+. All classes are well recognized except \emph{Fear} and \emph{Sad} are easily misclassified to \emph{Happy} and \emph{Angry}. (b): confusion matrix of our multi-scale model with our method on Oulu-CASIA databases. \emph{Happy} and \emph{Surprise} are recognized well while \emph{Angry} is easily to be confused with \emph{Disgust} symmetrically. Percent sign (\%) and cells with a rate less than 0.5\% are omitted for a better vision effect.}
\label{cm_ckp_oulu}
\end{figure}

\begin{figure}[tp]
\centering
\begin{minipage}{0.49\linewidth}
\includegraphics[width=\linewidth]{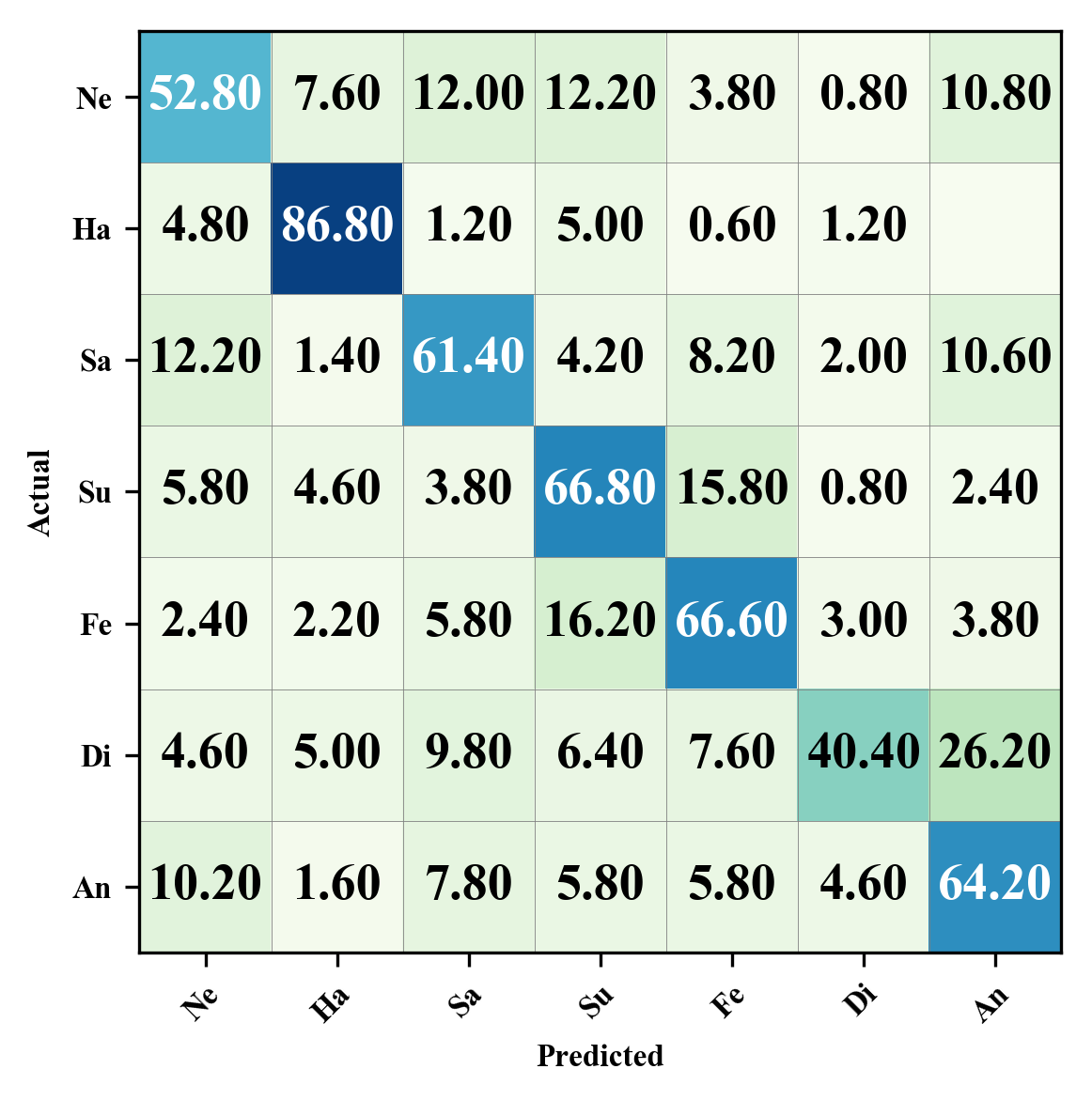}
\centerline{(a) with Ours}
\end{minipage}
\begin{minipage}{0.49\linewidth}
\includegraphics[width=\linewidth]{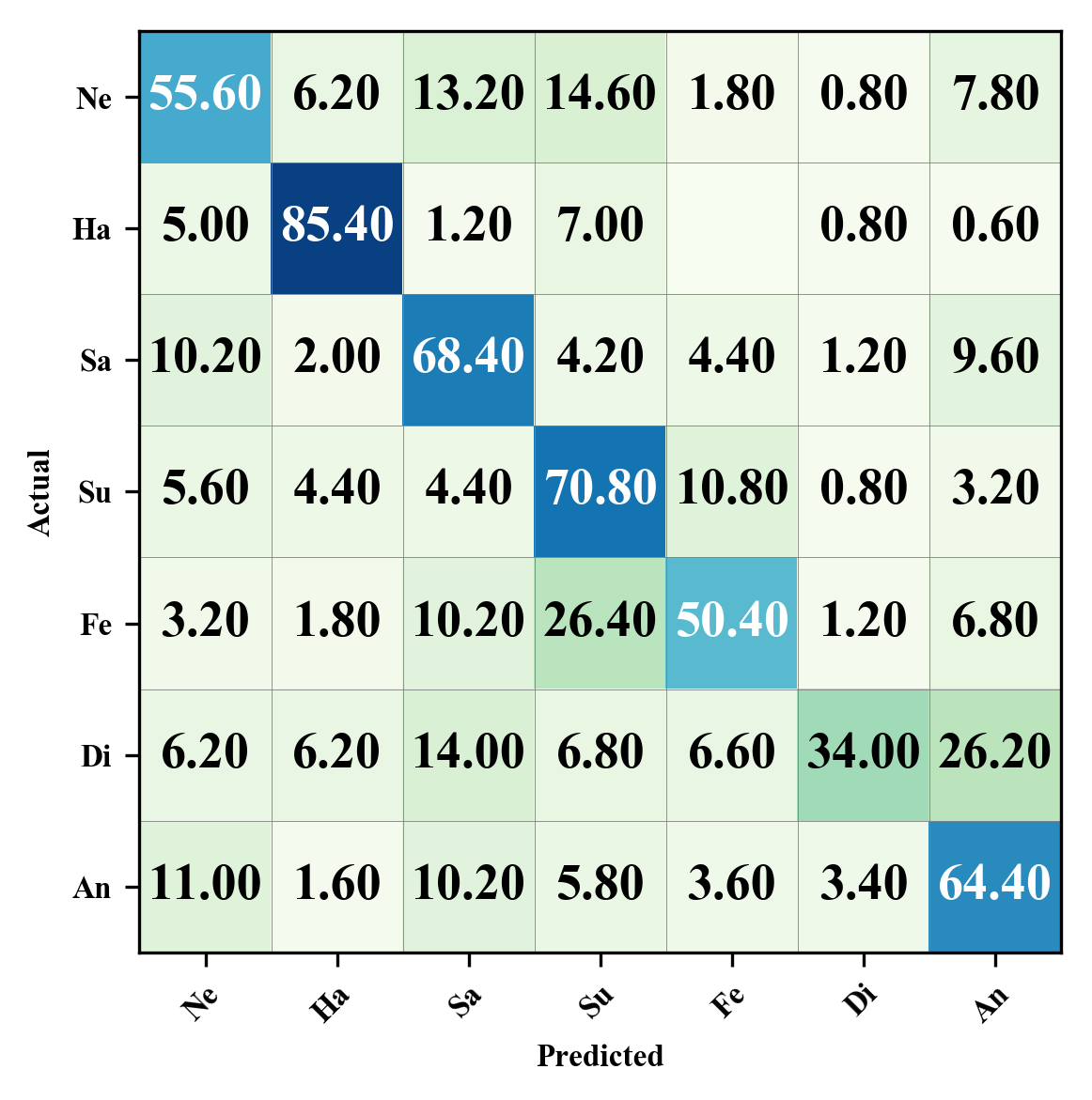}
\centerline{(b) w/o Ours}
\end{minipage}
\caption{
(a): confusion matrix of our model with our method on AffectNet. (b): confusion matrix of our baseline multi-scale model on AffectNet. Minority classes, i.e. \emph{Disgust} and \emph{Fear}, are predicted better in the former. Percent sign (\%) and cells with a rate less than 0.5\% are omitted for a better vision effect.}
\label{cm_aff}
\end{figure}

\begin{table}[tp]
\caption{Expression recognition average accuracy of different methods on Oulu-CASIA database. (\textbf{Bold}: Best result. \underline{Underline}: Second best.)}
\label{oulu_score}
\begin{center}
    \begin{tabular}{|c|c|}
    \hline
    \textbf{METHOD}&\textbf{ AVERAGE ACCURACY}\\
    \hline
    AdaLBP\cite{adalbp}&73.54\%\\
    Atlases\cite{atlases}&75.52\%\\
    STM-ExpLet\cite{stm}&74.59\%\\
    RADAP\cite{radap}&75.83\%\\
    FMPN\cite{fmpn}&86.33\%\\
    DTAGN\cite{prior-landmark2}&81.46\%\\
    LOMO\cite{lomo}&82.10\%\\
    FN2EN\cite{trans-f2e}&\underline{87.71\%}\\
    PPDN\cite{trans-peak}&84.59\%\\
    DeRL\cite{gen-id2}&\textbf{88.0\%}\\
    \hline
    \textbf{Our Method}&\textbf{87.6\%}\\
    \hline
    \end{tabular}
\end{center}
\end{table}

\noindent{\textbf{{Results on AffectNet:}}
TABLE~\ref{aff_score} reports the accuracy on AffectNet. Our multi-scale model shown in Fig~\ref{overview} can serve as a powerful baseline model and achieves an accuracy of 61.3\%, which is comparable to or even better than performance of other advanced method while the state-of-the-art accuracy is 61.5\%. With the help of our method, our model gains apparent performance improvement, getting closer to the upper limit, i.e. the agreement rate between annotators, than the state-of-the-art methods by a factor of 30\%. Also note that we use gray images of a size of $112\times112$ pixels while most of others produce the prediction under a resolution of $224\times224$ or higher.

Confusion matrices of models trained with/without our method are shown in Fig~\ref{cm_aff}. It can be noticed that the \emph{Happiness} emotion can be recognized well with an accuracy of 86.8\%, while the \emph{Disgust} expression achieves a lowest accuracy of 40.4\%. The enormous performance difference may be very owning to the extreme imbalance of classes distribution (48.3\% vs 1.3\%). Apart from that, compared with Fig~\ref{cm_aff}-(a), it can be observed that the recognition rates of minority classes, i.e. \emph{Disgust} and \emph{fear}, increases by 32.5\% and 18.8\% respectively, which means our method are more robust to the imbalance problem.

\noindent{\textbf{{Results on Oulu-CASIA and CK+:}}
TABLE~\ref{ckp_score} and TABLE~\ref{oulu_score} report the performance of our model on the two datasets, compared with other state-of-the-art methods. Our method achieves comparable accuracy with those advanced works with a narrow gap less than 1\%(only around one or less sample is misclassified during each testing).
Confusion matrices are illustrated in Fig~\ref{cm_ckp_oulu}. All classes are well recognized in CK+, except \emph{Fear} and \emph{Sadness}, which are easily misclassified to \emph{Happiness} and \emph{Anger}, respectively. That might due to the imbalance of classes contribution in CK+, where there are only 25 \emph{Fear} samples and 28 \emph{Sadness} samples in the whole dataset. In Oulu-CASIA, our model performs well on the emotion \emph{Happiness} and \emph{Surprise}, but is easy to confuse \emph{Anger} with \emph{Disgust} symmetrically.

\subsection{Ablation Study and Control Experiments}
\label{EXP_ablation}

\begin{table}[tp]
\caption{Expression recognition average accuracy of proposed method with simple threshold (T) or with batch-level threshold (BT), under different settings, on Affectnet database.(BS:Batch size. LR: Learning rate. Symbol $n\times$ means learning rate referred in this paper should be multiplied by $n$. A better performance under different batch size is in \textbf{Boldface})}
\label{aff_comp}
\begin{center}
    \begin{tabular}{|c|c|c|}
    \hline
    \textbf{METHOD}&\textbf{ SETTING}&\textbf{ ACCURACY}\\
    \hline
    \multirow{4}{*}{Ours(BT)}&BS:63 LR:$1\times$&\textbf{62.7\%}\\
    \cline{2-3}
    &BS:35 LR:$1/2\times$&61.5\%\\
    \cline{2-3}
    &BS:14 LR:$1/4\times$&\textbf{60.8\%}\\
    \cline{2-3}
    &BS: 7 LR:$1/8\times$&58.3\%\\
    \hline
    \multirow{4}{*}{Ours(T)}&BS:63 LR:$1\times$&62.1\%\\
    \cline{2-3}
    &BS:35 LR:$1/2\times$&\textbf{61.8\%}\\
    \cline{2-3}
    &BS:14 LR:$1/4\times$&60.6\%\\
    \cline{2-3}
    &BS: 7 LR:$1/8\times$&\textbf{59.9\%}\\
    \hline
    \end{tabular}
\end{center}
\end{table}

\noindent{\textbf{{Batch-level Threshold}} A comprehensive experiment comparing the proposed method with experimental threshold (noted by Ours(T)) or batch-level threshold (noted by Ours(BT)) is shown in TABLE~\ref{aff_comp}. When using a mini-batch size of 63, the batch-level threshold shows a best accuracy of 62.7\%, and the performance is then outperformed by the experimental threshold with the decreasing of mini-batch size. It is because that the threshold computed by averaging gets more unreliable when the size of mini-batch becomes relatively small, while in the other case the threshold is independent of mini-batch size.

\begin{table}[tp]
\caption{Results of validation on sequential datasets.(FC: Failure cases number. CFC: Consistent failure cases number.)}
\begin{center}
    \begin{tabular}{|c|c|c|c|}
    \hline
    \textbf{Dataset}&\textbf{SETTING}&\textbf{FC}&\textbf{CFC}\\
    \hline
    \multirow{2}{*}{\tabincell{c}{CK+}}
    &w/o Ours&15.4&0\\
    \cline{2-4}
    &with Ours&9&1.8\\
    \hline
    \multirow{2}{*}{\tabincell{c}{Oulu-CASIA}}
    &w/o Ours&14&0.2\\
    \cline{2-4}
    &with Ours&8.8&2.8\\
    \hline
    \end{tabular}
\end{center}
\label{sequence}
\end{table}

\noindent{\textbf{{Validation on Sequential Datasets:}} We also validate our method with real small facial variation of the same expression, designing an experiment to show the effect of our method. Sequential datasets, i.e. Oulu-CASIA and CK+ are used in experiment to show the stability of model output w.r.t. the small change of input images. As mentioned, every sequence of these two datasets records the change of expression, from neutral to peak, thus several near frames can be seen as \emph{a kind of} semantic neighbors and can be utilized for further validation. We train our model with the last, i.e. peak image, and test the model with other four images nearest to it.

We treat a group of those five images as a Failure Case (FC) when the prediction result of any test image in this group is inconsistent with the ground truth label. TABLE~\ref{sequence} records the average failure cases number of repeated test in 5 times, which decreases obviously with the usage of the proposed method. Apart from that, for further discussion, we also record the average number of Consistent Failure Cases (CFC), which refers to the groups where all test images are misclassified to a same expression.

Fig~\ref{case} shows several CFC. Cases in first three lines are from Oulu-CASIA, all of which are annotated with label \emph{Fear}. It can be noticed that the shown expressions are very different from other typical \emph{Fear} expressions especially in the second and third lines. Therefore, those samples tend to be treated as unreliable samples and get masked by our method, contributing little gradient when training, and are all wrongly predicted by our model.

CFC detected in CK+ are shown in the last two lines. While the case in the last line are the same with that in Oulu-CASIA, situation in penultimate line is more interesting to discuss. It can be seen that the subject in this clip gives the expression quite fast, keeping a neutral face until the last frame. Model training with our method fails in these four samples. However, when our method is not conducted, a well-trained model can still classify the four neutral face to \emph{Surprise} expression, which contrasts with our intuition and expectation. It exposes that model trained without our method may try to overfit samples(recognizing expressions by identity), and our method does remedy this tendency.
\begin{figure}[tp]
\centering
\includegraphics[width=\linewidth]{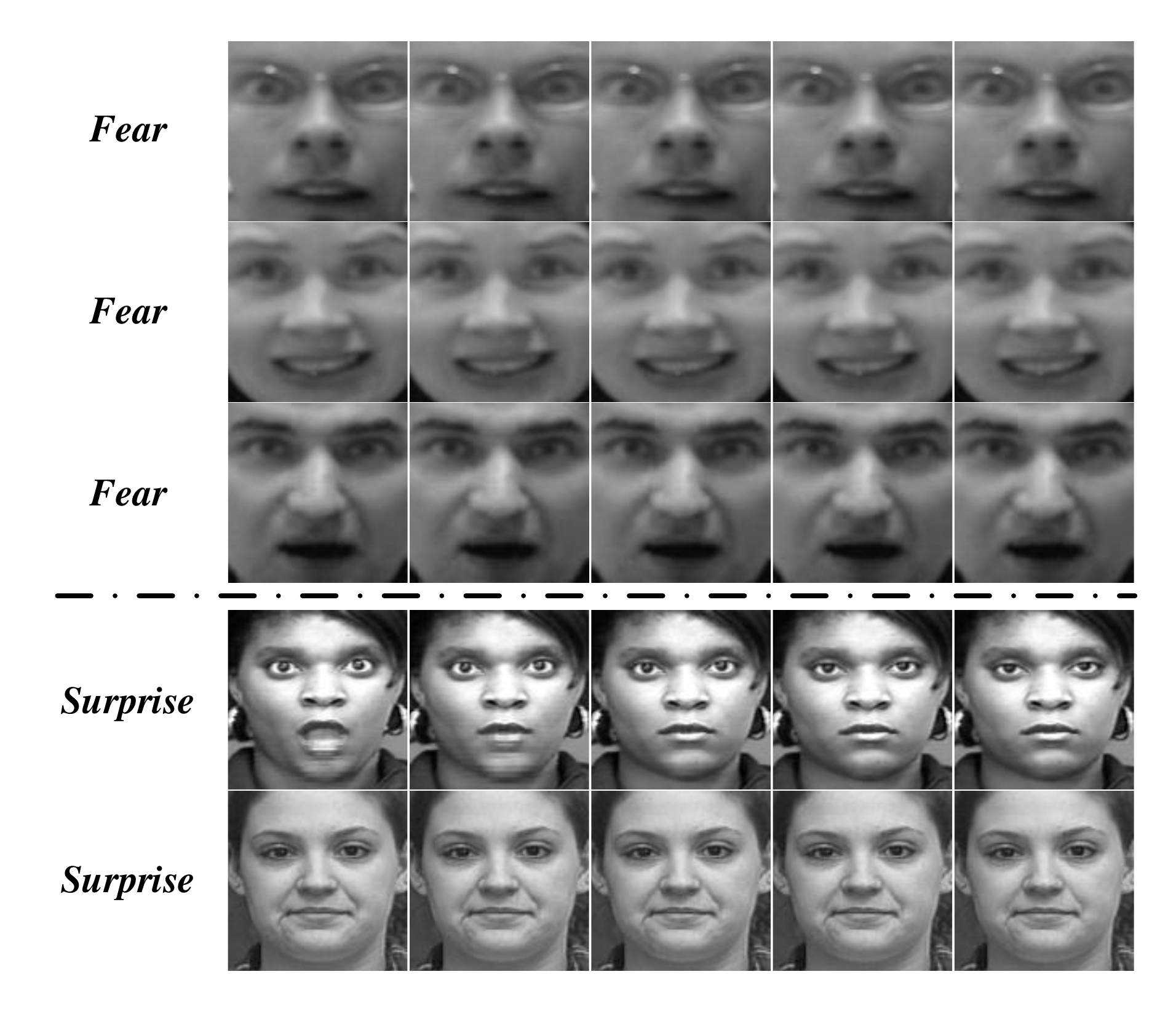}
\caption{
Consistent failure cases indicated by our method, from Oulu-CASIA (the first three lines) and CK+ (the last two lines).}
\label{case}
\end{figure}

\noindent{\textbf{{Ablation Study:}} To show the effectiveness of each optional setting, we do ablation study mainly on the options of \emph{Loss Function}, \emph{Balance Strategy}, and \emph{Our Method}. In addition, for a more clear and pointed view, when testing the effect of one optional setting we select the best choice in other optional setting groups except in the last one. Specifically, we evaluate the effectiveness of different loss functions under the setting of batch-balancing in \emph{Balance Strategy} and w/o Ours in \emph{Our Method}. Detailed results can be seen in TABLE~\ref{ablation}.

It can be observed that in this classification task, the usage of focal-loss has a little effect on the accuracy compared with cross-entropy. In this experiment \emph{Balance Method} does matter because classes distribution of this dataset is extremely imbalanced and the model will not achieve an acceptable result without any balance approaches. Weighted-loss approach leads to a significant performance gain and the batch-level-balancing then even improves the performance based on the former, which is comparable with the state of the art method. As for our proposed method, our method with a simple threshold gains apparent performance improvement, getting closer to the upper limit (65.3\%, rate of annotators agreement) than w/o Ours by a factor of 20\%, and the factor becomes 35\% in the case of our method with batch-level threshold.
\begin{table}[tp]

\caption{Results of ablation study. Control experiments are conducted to show the effectiveness of each optional setting. (BB: Batch balancing. FL: Focal loss. \textbf{Bold}: Best result. \underline{Underline}: Second best.) }
\label{ablation}
\begin{center}
    \begin{tabular}{|c|c|c|}
    \hline
    \textbf{OPTION}&\textbf{SETTING}&\textbf{ ACCURACY}\\
    \hline
    \multirow{2}{*}{\tabincell{c}{Loss Function\\\emph{(BB, w/o  Ours)}}}&Cross Entropy&61.1\%\\
    \cline{2-3}
    &Focal Loss&61.3\%\\
    \hline
    \multirow{3}{*}{\tabincell{c}{Balance Strategy\\\emph{(FL, w/o  Ours)}}}&No Balancing&44.4\%\\
    \cline{2-3}
    &Weighted Loss&60.8\%\\
    \cline{2-3}
    &Batch Balancing&61.3\%\\
    \hline
    \multirow{3}{*}{\tabincell{c}{Our Method\\\emph{(FL, BB)}}}&w/o  Ours&61.3\%\\
    \cline{2-3}
    &with Ours(T)&\underline{62.1}\%\\
    \cline{2-3}
    &with Ours(BT)&\textbf{62.7}\%\\
    \hline
    \end{tabular}
\end{center}
\end{table}

\section{CONCLUSION}\label{CON}
In this paper, we discuss the consistency between the scale of input semantic perturbation and of the output probability fluctuation to solve the contrary between FER task and recent deep learning fashion. An asymmetric autoencoder is designed to synthesize a semantic neighbor, and a novel semantic neighborhood-aware optimization method is proposed to reduce the stimulation from unreliable samples. A state-of-the-art performance is reported to prove the effectiveness of the proposed method. Our proposed method is compatible with other different models and training strategies, and requires no more computing power in testing. In the future, we may attempt to extend the semantic neighborhood to a global scope, or a transformation on feature level may be conducted to avoid the usage of complex image synthesis.


\ifCLASSOPTIONcaptionsoff
  \newpage
\fi



\bibliographystyle{IEEEtran}

\bibliography{TIP-21436-2019}{}
%



%
\vspace{-2.0\baselineskip}
\begin{IEEEbiography}[{\includegraphics[width=1in,height=1.25in,clip,keepaspectratio]{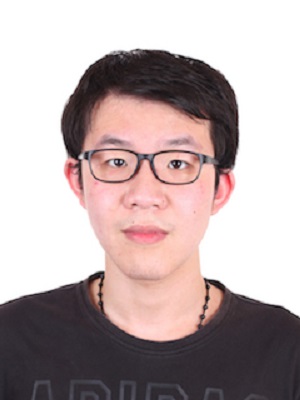}}]{Yongjian Fu}
received the B.S. degree from school of Computer Science and Technology, Wuhan University, Wuhan, China, in 2018. He is currently a PhD student in College of Computer Science at Zhengjiang University, China. His advisors are Prof. Zhijie Pan and Prof. Xi Li. His current research interests are primarily in machine learning and image processing.
\end{IEEEbiography}

\vspace{-1.5\baselineskip}
\begin{IEEEbiography}[{\includegraphics[width=1in,height=1.25in,clip,keepaspectratio]{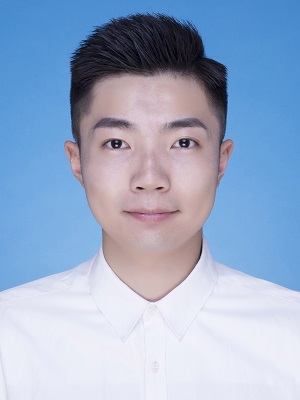}}]{Xintian Wu}
received the B.S. degree from school of Intelligent Science and Technology, Xidian University, Xi'an, China, in 2018. He is currently a PhD student in College of Computer Science at Zhengjiang University, China. His advisor is Prof. Xi Li. His current research interests are primarily in machine learning and image processing.
\end{IEEEbiography}

\vspace{-1.5\baselineskip}
\begin{IEEEbiography}[{\includegraphics[width=1in,height=1.25in,clip,keepaspectratio]{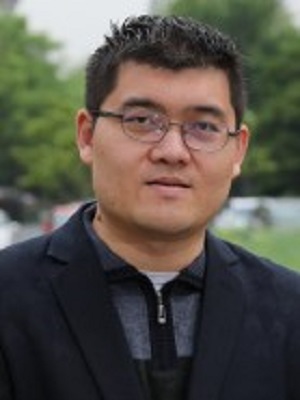}}]{Xi Li}
received the Ph.D. degree from the National Laboratory of Pattern Recognition, Chinese Academy of Sciences, Beijing, China, in 2009. From 2009 to 2010, he was a Post-Doctoral Researcher with CNRS, Telecomd ParisTech, France. He is currently a Full Professor with Zhejiang University, China. Prior to that, he was a Senior Researcher with the University of Adelaide, Australia. His research interests include visual tracking, motion analysis, face recognition, web data mining, and image and video retrieval.
\end{IEEEbiography}

\vspace{-1.5\baselineskip}
\begin{IEEEbiography}[{\includegraphics[width=1in,height=1.25in,clip,keepaspectratio]{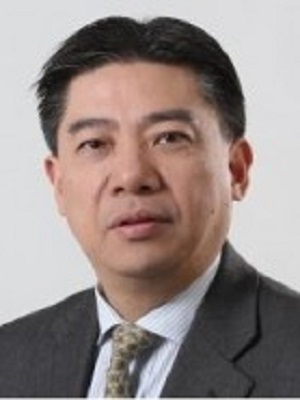}}]{Zhijie Pan}
is currently the Director of the Intelligent Vehicle Research Center, Zhejiang University. He is also mainly involved in the cross research fields of computer science and automotive, including AI, autonomous vehicle technology, automatic driving, and ITS. He has presented a series of new concepts in the fields of intelligent electric vehicle, chassis-control-by-wire technology, intelligent driving, unmanned systems, smart city, intelligent transportation, and applications. He has published more than 90 articles in international conferences and international magazines IEEE, SAE, and JSAE.
\end{IEEEbiography}

\vspace{-1.5\baselineskip}
\begin{IEEEbiography}[{\includegraphics[width=1in,height=1.25in,clip,keepaspectratio]{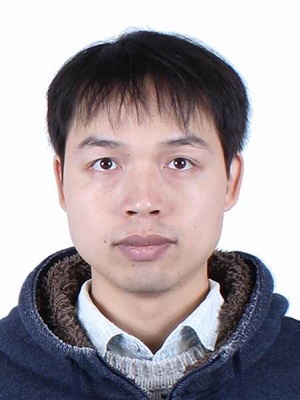}}]{Daxin Luo}
Dreceived the Ph.D. degree from the Institute of Semiconductors, Chinese Academy of Sciences, in 2013. He is currently a research engineer at Noah's Ark Lab, Huawei Technologies Co.Ltd. His research interests include computer vision and deep learning.
\end{IEEEbiography}





\end{document}